\pdfoutput=1 
\documentclass{article}

\usepackage[utf8]{inputenc} 
\usepackage[T1]{fontenc}    
\usepackage[colorlinks,citecolor=blue!70!black,linkcolor=blue!70!black,
urlcolor=blue!70!black]{hyperref}       
\usepackage{url}            
\usepackage{booktabs}       
\usepackage{amsfonts}       
\usepackage{nicefrac}       
\usepackage{microtype}      
\usepackage{xcolor}         
\usepackage{wrapfig}
\usepackage{sidecap}
\usepackage{lipsum}
\usepackage{fullpage}
\usepackage{graphicx}
\usepackage{subfigure}
\usepackage{booktabs} 
\usepackage{mkolar_definitions}
\usepackage{enumitem}
\usepackage{comment}

\usepackage{multirow}

\allowdisplaybreaks
\usepackage{natbib}

\usepackage{xspace}
\usepackage{amsthm,amsmath,amssymb}
\usepackage{amsfonts,dsfont}
\usepackage{mathtools}
\usepackage{tikz}
\usepackage{comment} 
\usepackage{wrapfig}
\usepackage[final]{neurips_2024}

\usepackage{caption}
\usepackage{parskip}
\usepackage{wrapfig}

\title{A Study of Plasticity Loss in \\On-Policy Deep Reinforcement Learning}

\author{
  Arthur Juliani \qquad Jordan T. Ash\\
  Microsoft Research NYC\\
\texttt{\{ajuliani, ash.jordan\}@microsoft.com} 
}
\date{}

\begin{document}
\maketitle

\begin{abstract}
Continual learning with deep neural networks presents challenges distinct from both the fixed-dataset and convex continual learning regimes. One such challenge is \textit{plasticity loss}, wherein a neural network trained in an online fashion displays a degraded ability to fit new tasks. This problem has been extensively studied in both supervised learning and off-policy reinforcement learning (RL), where a number of remedies have been proposed. Still, plasticity loss has received less attention in the on-policy deep RL setting. Here we perform an extensive set of experiments examining plasticity loss and a variety of mitigation methods in on-policy deep RL. We demonstrate that plasticity loss is pervasive under domain shift in this regime, and that a number of methods developed to resolve it in other settings fail, sometimes even performing worse than applying no intervention at all. In contrast, we find that a class of ``regenerative'' methods are able to consistently mitigate plasticity loss in a variety of contexts, including in gridworld tasks and more challenging environments like Montezuma's Revenge and ProcGen.
\end{abstract}

\section{Introduction}

In many important machine learning domains, such as continual learning and online reinforcement learning (RL), training data are not wholly available simultaneously. Rather, in these settings, data arrives sequentially over a long period of time. Irrespective of the available quantity of training data, we would typically like to have the highest performing model possible at any given point. Unfortunately, these sequential learning scenarios are known to present optimization issues for neural networks. Specifically, if a model is naively updated to convergence each time new data arrives, and the amount of training data effectively increases, the resulting model generalizes worse than it would under more standard, non-sequential constraints. This phenomenon has been introduced by~\citet{ash2020warm} as the ``warm-start problem,'' where it was found that randomly initialized models generalize well but are expensive to fit, and warm-started models generalize poorly but converge far more quickly.\looseness=-1


An adjacent problem to warm-starting is that of \textit{plasticity loss}~\citep{lyle2023understanding}, which presents itself both in the supervised learning and reinforcement learning settings. Like in the warm-start problem, plasticity loss describes a degradation in a model's ability to fit new data as training progresses, ultimately harming both sample efficiency and asymptotic performance. If an agent achieves worse performance when fitting first a source distribution and then a distinct target distribution than it would if instead trained on the target distribution in isolation, the degradation is attributed to plasticity loss.\looseness=-1

Whereas the warm-start problem is restricted to test performance, plasticity loss is generally measured as a degradation in performance on training data, often under some form of distribution shift. Despite the typical focus on training data, an ideal solution to plasticity loss would be able to prevent performance degradation both for data seen and unseen when fitting the policy. Importantly, the plasticity loss phenomenon is distinct from overfitting, as resolving plasticity loss should improve both training performance and generalization. In contrast, overfitting describes the opposite, where improving training performance has deleterious effects on generalization. 

This work studies the plasticity loss phenomenon in detail for the on-policy reinforcement learning setting. We introduce three distinct kinds of distribution shift to facilitate our analysis as well as a variety of environments and tasks. While several interventions have already been proposed, we demonstrate that some of the methods which proved successful for addressing plasticity loss in the supervised learning or off-policy reinforcement learning setting fail in the on-policy setting or under our proposed forms of distribution shift. We further highlight several criteria that we believe are necessary to remedy the issue. Based on these insights, we describe several techniques that resolve plasticity loss on the environments we consider.

In summary, the contributions of this paper are:

\begin{itemize}
    \item We extend studies of plasticity loss and the warm-start problem to the on-policy regime, finding that they present a persistent issue across a variety of model architectures and environmental distribution shift conditions.
    \item We provide an in-depth analysis of the correlates of these pathologies, studying several types of environmental settings, model architectures, and previously proposed approaches for mitigation in settings related to on-policy reinforcement learning. Importantly, we include generalization trends in our consideration of these phenomena.
    \item We characterize properties of methods that seem necessary for interventions to be successful at both addressing plasticity loss and ensuring generalization performance is maintained, and provide recommendations along these lines.
    
\end{itemize}

\section{Preliminaries}

An RL algorithm is described as ``on policy'' if it is trained using data collected from its own policy. While generally more sample inefficient than off-policy alternatives, where the data collection mechanism deviates from the policy being fit, on-policy methods have become more commonly used in practical RL problems because of their simplicity and stability~\citep{andrychowicz2020matters}.

Proximal Policy Optimization (PPO) is a ubiquitous on-policy deep RL algorithm~\citep{schulman2017proximal}. As a trust-region method, PPO constrains the extent by which a weight update can modify the current policy, quelling instability issues caused by high-variance policy gradients. Despite this robustness, on-policy algorithms still update on data corresponding only to the agent’s current experience, which may be sub-optimal for capturing necessary information around solving the underlying MDP. There is some evidence that plasticity loss can be present in on-policy RL learning problems~\citep{dohare2023maintaining}, demonstrating the effect in a somewhat ad hoc manner, we systematically use multiple environments and distribution shift conditions to more thoroughly characterize plasticity loss---and how to resolve it---in the on-policy setting.

\subsection{Simulating environmental distribution shift}

This paper primarily studies three distinct forms of distribution shift that we can reasonably expect a plasticity-preserving intervention to mitigate. Each experiment is organized into $r$ distinct rounds. In each round we supply $k$ environments from some distribution to the agent, fit a policy for that set of environments, and measure performance at the end of the round. To induce distribution shift, between each round we modify the data according to one of the three strategies described below. The environmental modifications we consider are: 

\begin{wrapfigure}{r}{0.5\textwidth}    
    \centering
    \vspace{-0.3cm}
    \includegraphics[width=1\linewidth]{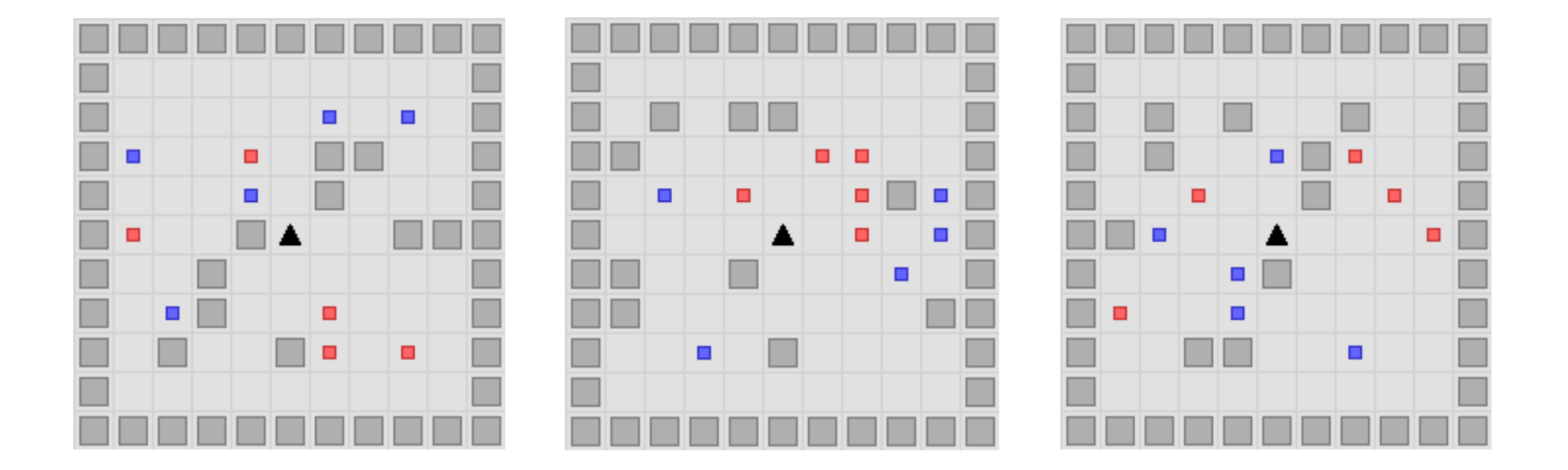}
    \caption{Examples of gridworld environment tasks. The agent (black triangle) begins each episode in the center of the environment. Blue jewels provide \textit{+1} reward, red jewels provide \textit{-1} reward, and dark grey walls prevent movement. Objects are placed randomly.\vspace{-0.3cm}}
    \label{fig:gridworld_example}
\end{wrapfigure}

\textbf{Permute.} Input pixels are randomly shuffled from round to round. Each of the $k$ environments in a given round are shuffled in the same way, but distinct from the shuffling in any other round. This method was utilized in~\citet{dohare2023maintaining}. We expect that an optimal learning algorithm will be able to achieve equivalent performance on the initial and subsequent rounds.

\textbf{Window.} We supply the agent with $k$ new environments for the same task. Here previously seen environments are no longer accessible, so the agent can only optimize with respect to the current $k$. This approach was considered by\\ \citet{abbas2023loss} and~\citet{lyle2023understanding}. We expect that an optimal learning algorithm will be able to achieve equivalent or better performance on subsequent rounds as compared to the initial round.

\textbf{Expand.} Like in the window modification, $k$ new environments are supplied to the agent. Here, however, they are appended to the total amount of training data, such that the number of available environments in the final round of training is $k \times n$. This is most similar to the warm-start environment studied primarily in~\citet{ash2020warm}, and revisited in~\citet{igl2020transient}. In this setting we expect training performance to decrease each round, even for an optimal learning algorithm, as available training data grows and becomes more difficult to fit. Correspondingly, we expect the generalization performance to increase as the training distribution expands to better capture the true distribution.

In the case of the latter two modifications, data at each new round are drawn from the exact same distribution as environments from previous rounds. As we demonstrate in Section~\ref{sec:plasticity_loss}, these still induce significant performance depredation in warm-started models. This observation is surprising---we often think of a model's initialization as a sort of ``prior'' over learned functions, and here we demonstrate that even an initialization obtained by training on data distributed identically to the current round still hinders performance in on-policy learning.

We use a simple gridworld task as our main sandbox for studying plasticity loss. The environment is drawn from the NeuroNav library~\citep{juliani2022neuro}, and the goal of the agent is to collect rewarding blue jewels while avoiding punishing red jewels in a fixed time window (100 time-steps per episode). Each sample of the environment from the distribution of possible tasks changes both the location of the jewels and the walls of the maze. Figure \ref{fig:gridworld_example} shows environment modification examples.

\section{Existing Approaches}\label{sec:methods}

A number of methods have been proposed to address plasticity loss in deep neural networks, though none were designed with on-policy RL explicitly in mind. Instead, these methods have been primarily validated in the context of either supervised learning or off-policy RL. Each of these techniques can be divided into one of two groups: Interventions which are performed intermittently during training, and those which are applied continuously, either as part of the model architecture or as part of the model update process. For additional implementation details see Appendix Section~\ref{app:method_details}.

\subsection{Intermittent Interventions}

We consider an intervention to be \textit{intermittent} if it is applied only at specific points during training. Periodically, the interventions in this category are applied, and training proceeds normally otherwise. Most of these are applied each time there is a training distribution change, implying that there must be awareness of when this occurs; in practice this information may be unavailable and difficult to detect.\looseness=-1

\textbf{Resetting final layer.} Proposed by \citeauthor{nikishin2022primacy}, this method involves periodically replacing the weights of the final layer in the network with newly initialized values. It was demonstrated that this alleviates plasticity loss in certain off-policy RL settings.

\textbf{Shrink+Perturb.} This technique periodically scales the magnitude of all weights in the network by a factor and then adds a small amount of noise~\citep{ash2020warm}. It was demonstrated that this improves performance in batched continual learning settings. In our implementation the weight of these two factors are entangled such that they sum to one.

\textbf{Plasticity Injection.} Plasticity Injection replaces the final layer of a network with a new function that is a sum of the final layer's output and the output of a newly initialized layer subtracted by itself~\citep{nikishin2023deep}. The gradient is then blocked in both the original layer and the subtracted new layer. It was shown that this method increases performance of off-policy RL agents in the ALE.

\textbf{ReDo.} This technique resets individual neurons within the network based on a dormancy criteria at fixed intervals~\citep{sokar2023dormant}, leading to improvement in performance in the context of off-policy RL agents trained on the ALE. Following \citeauthor{sokar2023dormant}, we reset the model parameters more frequently than we apply environmental distribution shifts.

\subsection{Continuous Interventions}

We characterize a method as \textit{continuous} if it is applied at every step of optimization. Continuous interventions are desirable because they do not require an awareness of when a distribution shift has occurred, which can be challenging in practical scenarios where we want to mitigate plasticity loss.

\textbf{L2 Norm.} Discussed by \citeauthor{lyle2023understanding}, this method involves regularizing the network using the L2 norm. It was demonstrated that in certain continual learning settings this reduces plasticity loss.

\textbf{LayerNorm.} LayerNorm~\citep{ba2016layer} is a ubiquitous deep learning regularization technique, and was shown useful for mitigating plasticity loss by \citeauthor{lyle2023understanding}. The authors demonstrated a performance improvement in the ALE when using off-policy RL.

\textbf{CReLU activations.} Proposed by \citeauthor{abbas2023loss} and studied further by \citeauthor{lee2024plastic}, this method involves replacing ReLU activations with the CReLU activation function. This ensures that the gradient is non-zero for all units in a given layer, which appears to mitigate plasticity loss when periodically switching between tasks in the ALE with off-policy methods.

\textbf{Regenerative Regularization.} Regenerative regularization uses an L2 penalty to encourage model weights to be near their initial values \citep{kumar2023maintaining}. It has been shown to be effective in a number of continual learning settings. 

\vspace{-0.1cm}
\section{Experiments}\label{sec:experiments}
\vspace{-0.1cm}
\begin{figure*}[!t]
    \centering
    \includegraphics[width=1.0\linewidth]{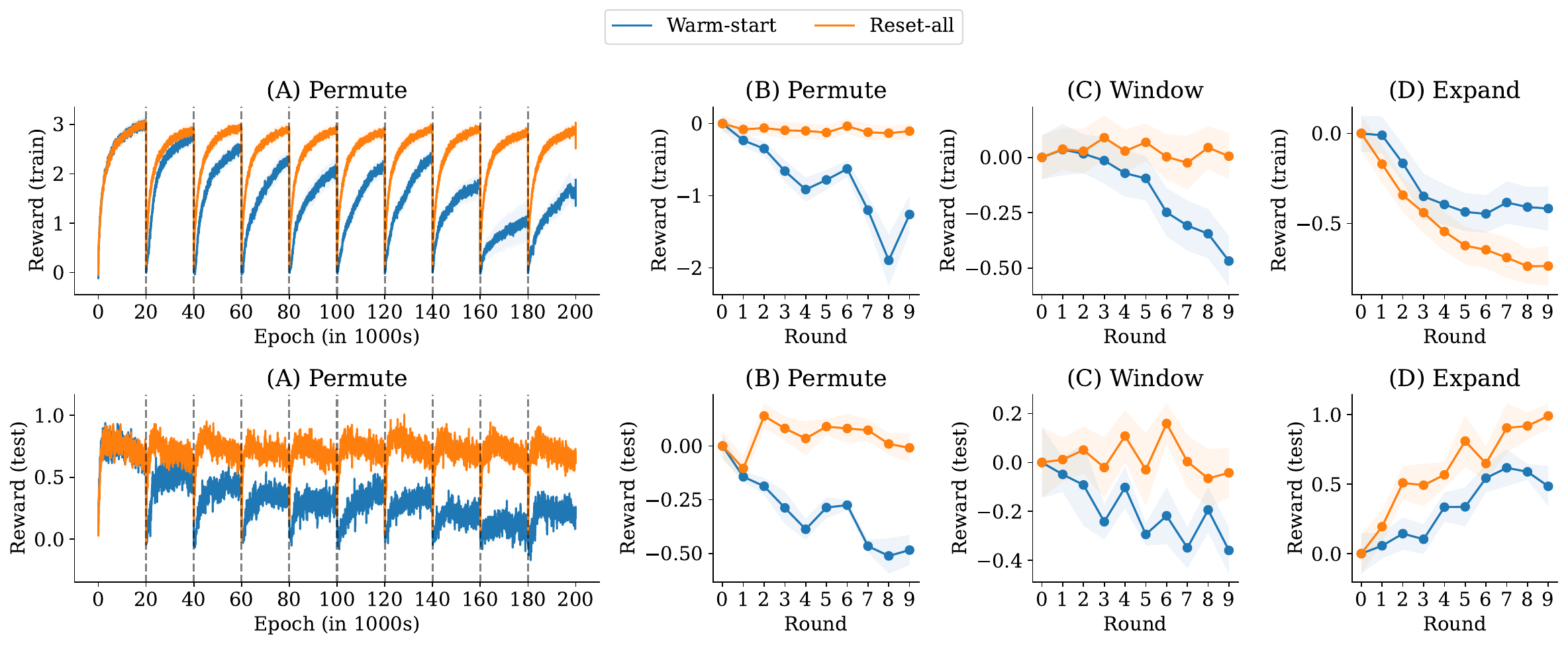}
    \caption{Performance in the gridworld environment under each of the three distribution shift conditions. The degradation between rounds is evidence of plasticity loss. (A): Epoch-level training performance for permute modification. Dotted vertical lines indicate the end of each round, before a new environmental distribution shift is applied. (B): Round-level training performance for the Permute modification. Data points correspond to normalized mean reward in final 50 episodes of round. Shaded regions correspond to standard error. (C, D): Round-level training performance for the Window and Expand conditions. \textbf{Top row}: Training performance. \textbf{Bottom row}: Test performance. \vspace{-0.5cm}}
    \label{fig:plasticity_loss}
\end{figure*}

Below we present results of a suite of experiments conducted using a 2D gridworld environment~\citep{juliani2022neuro}, the procedurally generated CoinRun environment~\citep{cobbe2019quantifying}, and the Atari game Montezuma's Revenge~\citep{bellemare2013arcade}. All models are trained using PPO~\citep{schulman2017proximal}. For the gridworld experiments we use an MLP encoder, and for the CoinRun and Montezuma's Revenge experiments we instead use a convolutional encoder. We set the number of environment instances ($k$) to 100 for all experiments conducted with the gridworld and CoinRun environments. Additional training details can be found in Appendix Section~\ref{app:env_details}.

\subsection{Plasticity loss is apparent in on-policy RL}\label{sec:plasticity_loss}

We find evidence of plasticity loss in the on-policy RL setting in the presence of multiple kinds of distribution shift, as presented in Figure~\ref{fig:plasticity_loss}. For round-level plots like these, reward values are normalized such that the mean reward at the end of the first round is set to zero. As such, positive and negative values correspond respectively to increases and decreases in performance relative to the first round. Decreasing normalized reward indicates a decay in relative performance, implying plasticity loss.\looseness=-1

For both the permute and window shift conditions, Figure~\ref{fig:plasticity_loss} demonstrates clear degradation in mean episodic reward for the warm-started model as compared to a model which has all of its weights reset between each round. While there is some positive transfer for the warm-start method in the expand shift condition, we find generalization performance suffers in this condition as compared to reset-all. We provide statistical tests comparing the performance of these methods to the warm-start and random initialization baselines in Section \ref{app:results} of the appendix. 

\subsection{Predictors of plasticity loss and generalization}

\begin{figure*}[!t]
    \centering
    \includegraphics[width=1.0\linewidth]{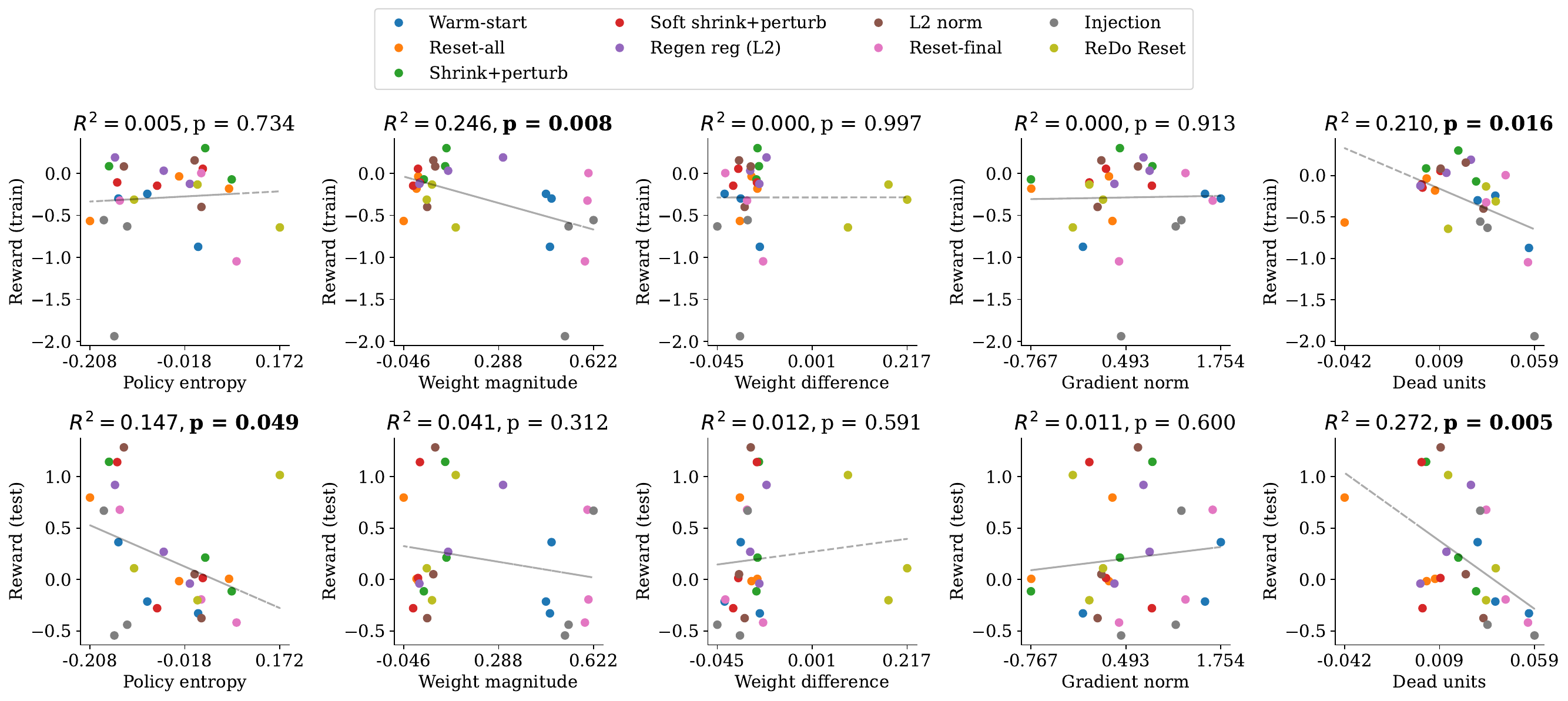}
    \vspace{0.1cm}
    \includegraphics[width=1.0\linewidth]{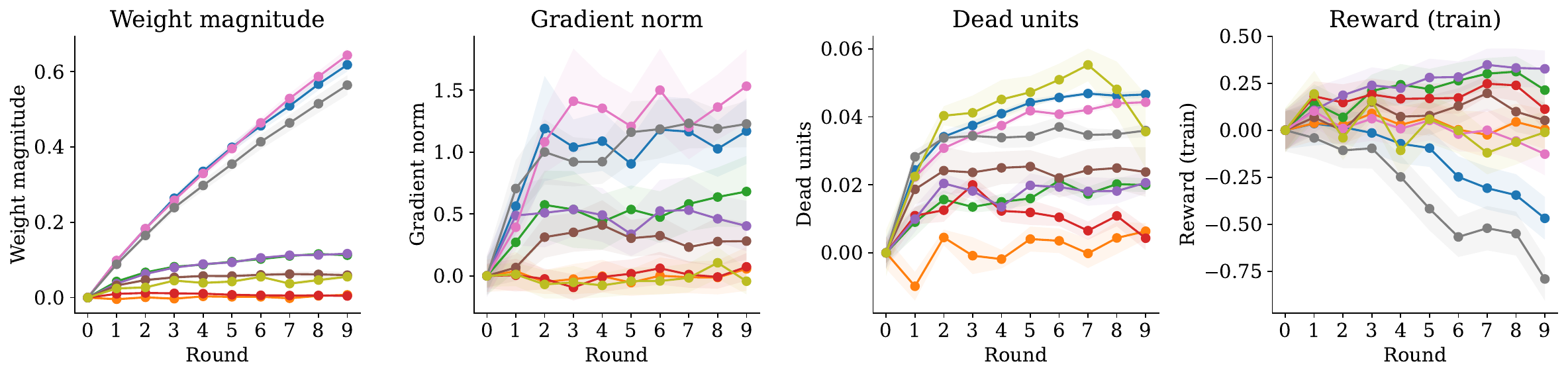}
    \caption{\textbf{Top}: Correlation plots of normalized mean reward in the gridworld environment compared against identified metrics. Each point is averaged over five replicates, and shows the final values produced after a ten-round experiment.
    Values for each measurement are normalized by its baseline level at initialization. Measurements that significantly correlate ($p < 0.05$) with normalized reward are bolded. \textbf{First row}: Training distribution performance. \textbf{Second row}: Test distribution performance. \textbf{Bottom}: Values of the three predictive metrics during the course of training for each intervention and window change condition as compared to training performance. Shaded regions correspond to standard error. Final values of plots like these correspond to a single point in the correlation plots above.\looseness=-1}
    \label{fig:metrics}
\end{figure*}

Identifying the exact statistical underpinnings of plasticity loss is an active area of research in continual and reinforcement learning. Here we identify several factors which are significantly correlated with plasticity loss as well as a degradation in generalization performance. Figure \ref{fig:metrics} presents correlation plots of normalized reward with various metrics of interest. Here we normalize rewards (i.e. subtract the initial reward value from the final reward value) to compensate for the fact that many of these algorithms have a regularization effect that improves performance on average but does not necessarily improve performance trends—it is possible to improve generalization, for example, without remedying plasticity loss.

Note that the \textit{LayerNorm} and \textit{CReLU} methods are not included in this analysis, due to their direct effects on the weight magnitude and dead unit counts, respectively. For graphs of the underlying metrics over time during training, see Supplemental Figure \ref{fig:extra_metrics_full}. We find that weight magnitude and number of dead units are both significantly correlated with the normalized reward [$p < 0.05$]. ``Dead units'' refer to ReLU activations that never produce non-negative outputs.

We find that neither the gradient magnitude nor the magnitude of weight difference between updates has a significant, linear relationship with either plasticity or generalization. This is in contrast to the findings of \citeauthor{abbas2023loss} and \citeauthor{nikishin2022primacy}, each of which examined plasticity loss in off-policy rather on-policy settings. The former found that weight difference was correlated with loss of plasticity while the latter found that gradient norms were predictive.

We also fit a generalized linear model (GLM) using all five metrics as independent variables and normalized reward as the dependent variable. Here we find that both gradient and weight magnitudes are significantly predictive of train performance [$p < 0.01$], while dead unit count no longer is [$p > 0.05$]. This suggests that the predictive power of the dead unit count can be accounted for by the weight magnitude. An additional GLM was fit to the test data, and we find that only the dead unit count remains predictive [$p < 0.05$].

Although these analyses do not establish causality, the robust significance of weight magnitude in predicting plasticity loss suggests that plasticity loss may be a function of how much trainable parameters deviate from their initialization distribution. As such, successful mitigation strategies would induce a sub-linear increase in the weight norm of the network parameters over the course of learning. This has the downstream effect of minimizing the number of dead units in the network. Below we compare the mitigation strategies, finding that those which address this underlying cause directly indeed perform best.

\subsection{Novel architectural methods do not fully address plasticity loss}

\begin{figure*}[t]
    \centering
    \includegraphics[width=1.0\linewidth]{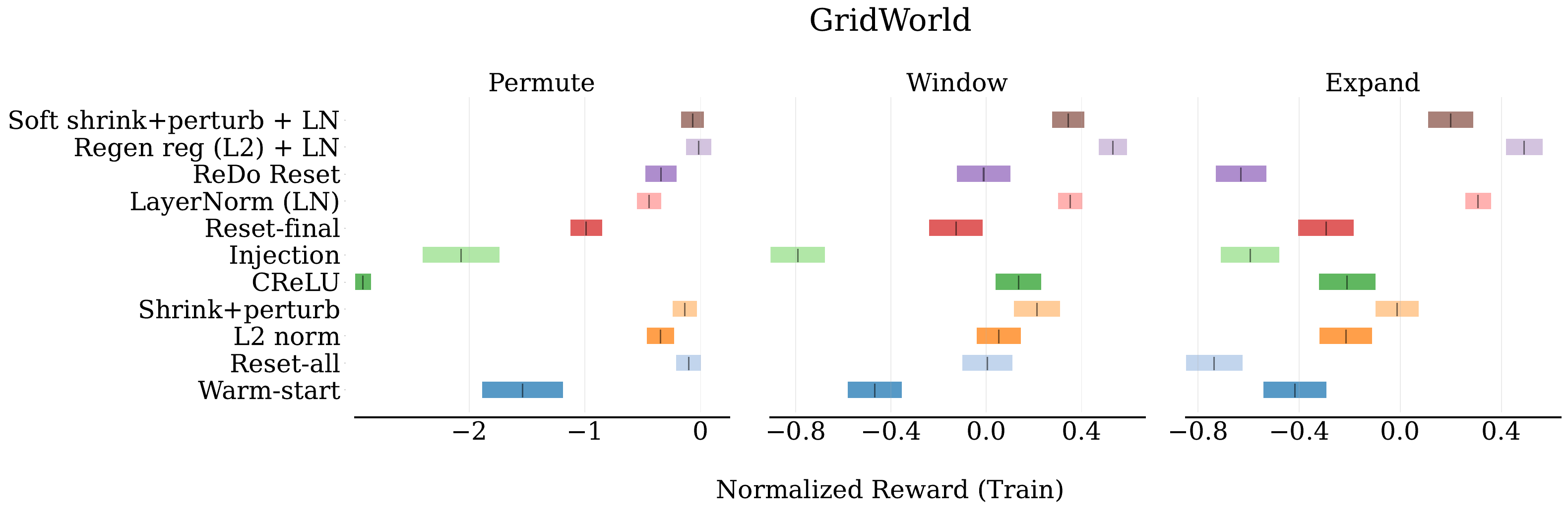}
    \includegraphics[width=1.0\linewidth]{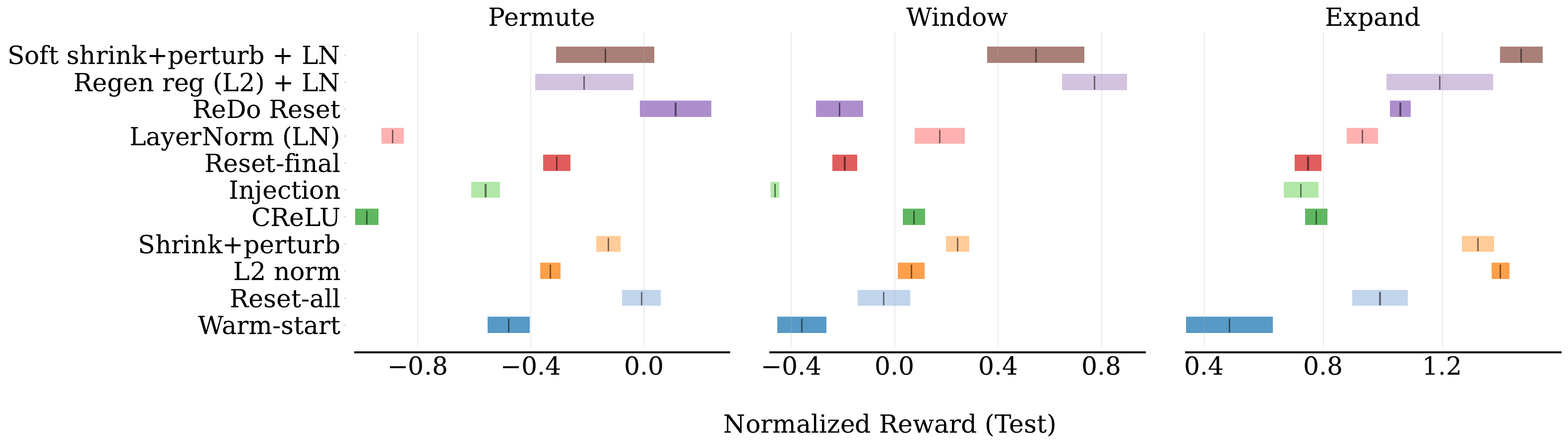}
    \caption{Performance of intervention methods compared to warm-start and reset-all baselines on the \textit{Gridworld} environment. Final round mean reward is normalized by the performance at end of the first round, and interval bars denote standard error. \textbf{Top}: Train performance. \textbf{Bottom}: Test performance. \looseness=-1}
    \label{fig:gridworld}
\end{figure*}

Recently a number of methods have been introduced to specifically address plasticity loss using changes to the network architecture. These methods were previously validated in the off-policy setting only. Here we consider the behavior of two of these methods as it applies to the on-policy setting: \textit{CReLU} \cite{abbas2023loss} and \textit{Plasticity injection} \cite{nikishin2023deep}. We find that both methods fail to address plasticity loss in the permute shift condition. The \textit{CReLU} method is able to address plasticity loss in the window and expand settings only, while the plasticity inject method underperforms even the warm-start baseline in all three contexts (Figure~\ref{fig:gridworld}).\looseness=-1 

\subsection{Regularization methods address plasticity loss}

We next consider the class of methods which regularize the weights of the network, either continuously or intermittently. These methods include \textit{final layer reset} \citep{nikishin2022primacy}, \textit{shrink+perturb} \citep{ash2020warm}, a continuous ``soft'' variant of shrink+perturb \citep{dohare2023maintaining}, \textit{L2 regularization}, \textit{ReDo} \citep{sokar2023dormant}, and \textit{regenerative regularization} \citep{kumar2023maintaining}. We find \textit{final layer reset} fails to address plasticity loss in any of the three conditions. In contrast, the other five methods mitigate both plasticity loss and the warm-start problem to some extent in all three shift conditions (Figure~\ref{fig:gridworld}). Of these methods, \textit{ReDo} performs the worst as it is unable to benefit from the positive transfer which is possible in the cases of the ``window'' and ``expand'' conditions, as methods that retain information from previous rounds effectively have access to more training data in these conditions. \textit{ReDo} underperforms even the warm-start baseline in the ``expand'' condition.\looseness=-1

\subsection{LayerNorm addresses training plasticity loss but not generalization}

\begin{figure*}[!t]
    \centering
    \includegraphics[width=1.0\linewidth]{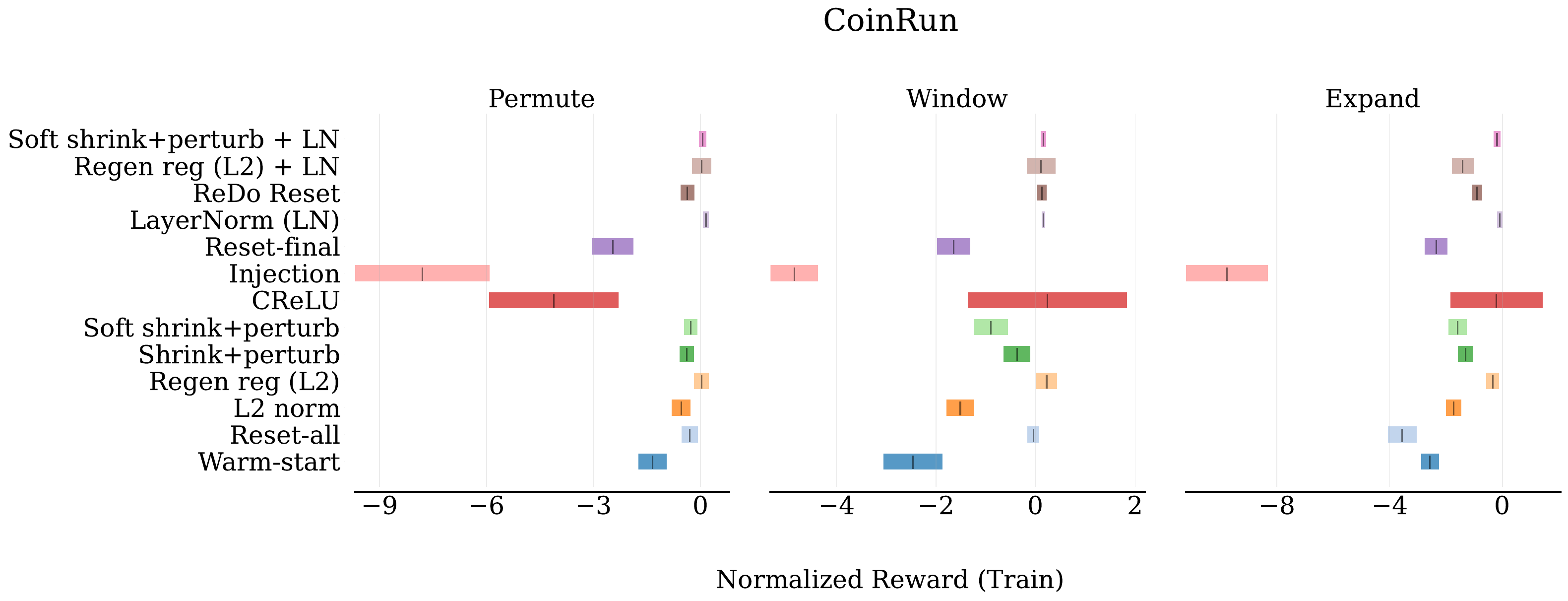}
    \includegraphics[width=1.0\linewidth]{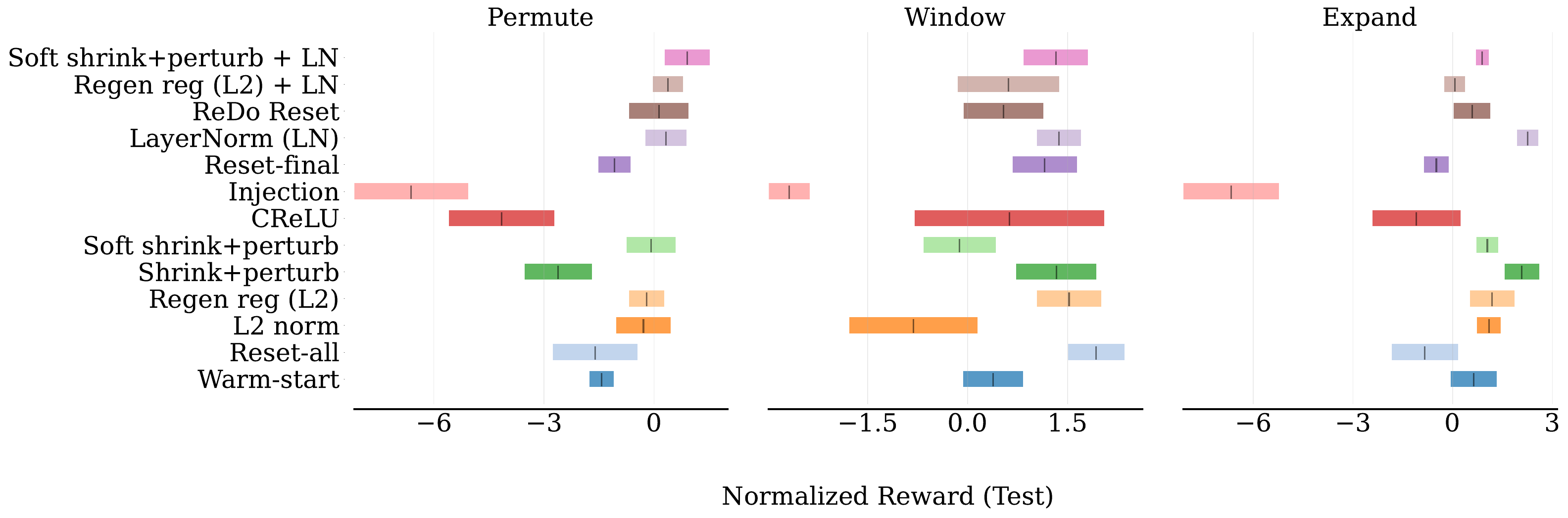}
    \caption{Performance of intervention methods compared to warm-start and reset-all baselines on the \textit{CoinRun} environment. Final round mean reward is normalized by the performance at end of the first round, and interval bars denote standard error. \textbf{Top}: Train performance. \textbf{Bottom}: Test performance. \looseness=-1}
    \label{fig:coinrun}
\end{figure*}

We also consider the effect of LayerNorm on plasticity loss. It was previously demonstrated in the off-policy setting that LayerNorm was able to mitigate some effects of plasticity loss~\citep{lyle2023understanding}. Given that LayerNorm tends to improve learning across a number of contexts~\citep{ba2016layer}, it is important to attempt to separate the effects of LayerNorm on performance generally from its effect on plasticity specifically. We find that LayerNorm resolves plasticity loss in terms of training performance, but is inconsistent in its effect on generalization performance. For example, the LayerNorm model significantly underperforms even the warm-start baseline on the test data of the permute shift condition. We find however that combining \textit{soft shrink+perturb} or \textit{regenerative regularization} with LayerNorm results in overcoming both training plasticity loss and deleterious generalization trends (Figure \ref{fig:gridworld}).

\subsection{Plasticity loss in \textit{CoinRun}}

We also use the CoinRun environment from the ProcGen suite of tasks to evaluate the set of candidate interventions \cite{cobbe2019quantifying,cobbe2020leveraging}. ProcGen is built using procedural generation \cite{cobbe2020leveraging}, making it possible to study the same three distribution-shift conditions which were considered in the gridworld experiments. Here we present performance results for each of the intervention methods on the CoinRun environment. 

Overall, we find results largely consistent with those of the gridworld task. There is significant plasticity loss present in all three tasks as measured by the gap between the warm-start and reset-all conditions. We also find that \textit{reset-final}, \textit{CReLU}, and \textit{plasticity injection} all unable to fully resolve the loss of plasticity. In contrast, the three most effective methods in the gridworld task (\textit{LayerNorm}, \textit{regenerative regularization}, and \textit{soft shrink+perturb}) are all also most effective in CoinRun as well. 

\begin{wrapfigure}{r}{0.5\textwidth}
    \centering
    \vspace{-0.0cm}
    \includegraphics[width=0.94\linewidth]{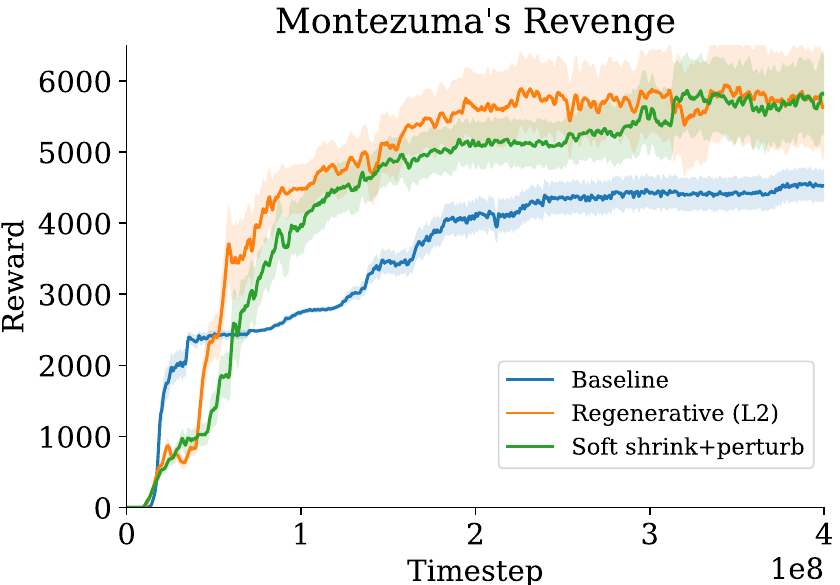}
    \caption{A comparison between RND agents trained with and without a plasticity-loss mitigating intervention. Experiments are done over twenty replicates and shaded regions show standard error.\vspace{-0.7cm}}
    \label{fig:montezuma}
\end{wrapfigure}

See Figure \ref{fig:coinrun} for plots of the normalized mean rewards over the course of training for all methods. For results on two additional ProcGen environments, see Appendix \ref{app:results}. For round-level performance plots of all methods, see Section \ref{app:results} in the appendix. We find that relative efficacy of the intervention methods is consistent across ProcGen environments.

\subsection{Plasticity loss in \textit{Montezuma's Revenge}}

Montezuma's Revenge is an Atari game that is often used to benchmark the quality of exploration procedures in RL~\cite{bellemare2013arcade,salimans2018learning}. The environment consists of many separate rooms, some only accessible through a locked door. The agent must avoid obstacles, find and use keys, and traverse several rooms before obtaining non-trivial reward. Because of this reward sparsity, exploration is needed to incentivize the agent to advance beyond the periphery of its experience. 

Montezuma's Revenge is designed such that the agent only sees the room it currently occupies. Accordingly, the distribution over input states expands as the agent learns to explore the environment. This introduction of significantly different states induces a level of distribution shift which is unique among ALE environments. Within our framework, the distribution shift of Montezuma's Revenge is most similar to the ``expand'' condition studied here. 

In this experiment we consider two versions of a policy trained using the Random Network Distillation (RND) exploration bonus~\citep{burda2018exploration}. All architectures and hyperparameters are inherited from the original RND paper. Figure~\ref{fig:montezuma} compares a typical RND policy to ones that are trained in conjunction with the \textit{soft shrink+perturb} or \textit{regenerative regularization} interventions. The plasticity loss mitigating policy achieves higher reward than a corresponding agent fitted in a standard fashion. Due to computation constraints, we did not evaluate the other intervention methods.

\section{Discussion}

Plasticity loss has been identified and rigorously studied in the continual learning and off-policy reinforcement learning settings. We find that it is also an issue in the on-policy setting under a variety of different conditions. Methods introduced in the off-policy setting such as \textit{plasticity injection} and \textit{CReLU} appear to be not as consistently effective in the on-policy setting. We hypothesize that this degradation may be due to the violation of the iid assumption which supervised learning and off-policy reinforcement learning rely on. Further, these methods require architectural modifications of the policy and value networks, and are thus less general than regularization-based alternatives.\looseness=-1

In contrast, the class of regularization methods, which include \textit{L2 regularization}, \textit{ReDo}, \textit{shrink+perturb}, and \textit{regenerative regularization} all are able to significantly mitigate plasticity loss in our experiments. Of these methods, \textit{soft shrink+perturb} displayed the best generalization performance, and can be easily combined with layer normalization. \textit{LayerNorm}, a now ubiquitous method in supervised learning, was previously introduced to address plasticity loss in the off-policy setting~\cite{lyle2023understanding}, and we find that it is also effective in the on-policy setting. Due to its general regularizing effects, \textit{LayerNorm} also increases baseline performance in the absence of distributional shift~\cite{ba2016layer}, suggesting that it is a generally useful method for on-policy reinforcement learning methods such as PPO.


A consistent feature of the class of regularization methods discussed here is that they all normalize the network parameters towards their initial distribution. This has the effect of decreasing the weight magnitude and reducing the number of dead or saturated activation units, as seen in Figure \ref{fig:metrics}. It has recently been proposed that both these metrics may be a proxy for the underlying curvature of the optimization landscape~\citep{lewandowski2023curvature}. Studying this connection more deeply would potentially be a fruitful avenue for future research.

\section{Related Work}

In the context of generalization in continual learning settings, a related issue to plasticity loss is the warm-start problem \citep{ash2020warm}. Recent work has shown that the benefits of methods such as shrink+perturb to generalization were dependent on the presence of noise in the labels used for training~\citep{zaidi2023does}. In the absence of noisy labels, other regularization methods were found to be competitive with methods that utilize re-initialization such as shrink+perturb.

Another perspective on the problem of loss of plasticity has focused on the content of early learning in a network. ~\citet{achille2018critical} demonstrate that neural networks have critical early periods of sensitivity to training data which are analogous to those in animals~\citep{hensch2004critical}. In the biological context this sensitivity is tied directly to neuronal plasticity. This connection to critical period learning was also drawn in work on off-policy RL, in which a ``primacy bias'' was proposed to explain the loss of plasticity seen in a number of task domains~\citep{nikishin2022primacy}. Utilizing this insight, the periodic soft-resetting of the neural network has enabled significant increases in the sample efficiency of off-policy RL methods~\citep{d2022sample,schwarzer2023bigger}.

In the deep RL domain a similar problem called \textit{ray interference} has been identified~\citep{schaul2019ray}. Ray interference arises from the conflicting gradient signals related to unique sub-tasks required to solve a single more complex task in online deep reinforcement learning. It is possible that for complex tasks such as CoinRun or Montezuma's Revenge there are unique sub-tasks which are learned and whose gradients may interfere with one another. Exploring the distinction between ray interference and plasticity loss in the RL setting is a promising future direction.

In the off-policy RL setting the problem of plasticity loss has also been studied under the name of ``capacity loss''~\citep{lyle2022understanding}. While plasticity loss refers to a property of the network-task interaction, capacity loss is a more generic term to refer to the properties of a network that are task invariant. This work advocates for a solution similar to that of~\citet{kumar2023maintaining}, and can thus be considered in the class of methods which regularize the network towards an initialization distribution.

Within the domain of on-policy RL there is a related phenomena to plasticity loss called ``policy collapse,'' which refers to a degradation of performance as training progresses, even on a fixed dataset~\citep{dohare2023overcoming}. Although these are distinct phenomena, there is likely overlap in the success of mitigation methods. A novel optimizer strategy ``Non-stationary Adam'' was proposed to address policy collapse, and although we did not study it, it may provide benefits on the tasks examined here. In terms of direct studies of plasticity loss in on-policy RL, both~\citet{dohare2023loss} and \citet{igl2020transient} show plasticity loss in the on-policy setting, and propose a method to address it.

As part of their analysis, \citet{dohare2023maintaining} propose a plasticity-enhancing method called ``Continual Backprop,'' which is another form of regularization towards the initialization distribution, however it is one which is selective on a per-neuron basis. This method is similar to ReDo~\citep{sokar2023dormant}, which we analyze here. It is also related to DrM~\citep{xu2023drm}, which builds on the DrQ algorithm.

It is finally worth mentioning that notions of plasticity loss have been studied in the neuroscience community as well. These range from classic work demonstrating the inability of animal models to fully accommodate new sensory input after developing cognitively without it~\citep{wiesel1963effects}, to more recent work characterizing plasticity loss in humans as a component of various psychopathologies~\citep{peled2005plasticity,carhart2023canalization,juliani2023deep}.

\section{Conclusion}

In this work we studied the problem of plasticity loss in the context of on-policy deep RL. We find that similar to continual learning and off-policy RL, plasticity loss is an issue across a number of different environments and forms of distributional shift. We find that some methods previously proposed to resolve the issue in other problem settings, such as \textit{CReLU} and \textit{plasticity injection} do not transfer to on-policy RL. The methods which best resolve plasticity loss are those which act as continual regularizers as opposed to intermittent interventions. Within this class of methods we find that a soft variant of \textit{shrink+perturb} combined with LayerNorm performs the best across our evaluated settings, providing a simple and general method for addressing plasticity loss.



\bibliography{bibliography}

\begin{thebibliography}{}

\bibitem[Abbas et~al., 2023]{abbas2023loss}
Abbas, Z., Zhao, R., Modayil, J., White, A., and Machado, M.~C. (2023).
\newblock Loss of plasticity in continual deep reinforcement learning.
\newblock {\em arXiv preprint arXiv:2303.07507}.

\bibitem[Achille et~al., 2018]{achille2018critical}
Achille, A., Rovere, M., and Soatto, S. (2018).
\newblock Critical learning periods in deep networks.
\newblock In {\em International Conference on Learning Representations}.

\bibitem[Andrychowicz et~al., 2020]{andrychowicz2020matters}
Andrychowicz, M., Raichuk, A., Sta{\'n}czyk, P., Orsini, M., Girgin, S., Marinier, R., Hussenot, L., Geist, M., Pietquin, O., Michalski, M., et~al. (2020).
\newblock What matters in on-policy reinforcement learning? a large-scale empirical study.
\newblock {\em arXiv preprint arXiv:2006.05990}.

\bibitem[Ash and Adams, 2020]{ash2020warm}
Ash, J. and Adams, R.~P. (2020).
\newblock On warm-starting neural network training.
\newblock {\em Advances in neural information processing systems}, 33:3884--3894.

\bibitem[Ba et~al., 2016]{ba2016layer}
Ba, J.~L., Kiros, J.~R., and Hinton, G.~E. (2016).
\newblock Layer normalization.
\newblock {\em arXiv preprint arXiv:1607.06450}.

\bibitem[Bellemare et~al., 2013]{bellemare2013arcade}
Bellemare, M.~G., Naddaf, Y., Veness, J., and Bowling, M. (2013).
\newblock The arcade learning environment: An evaluation platform for general agents.
\newblock {\em Journal of Artificial Intelligence Research}, 47:253--279.

\bibitem[Burda et~al., 2018]{burda2018exploration}
Burda, Y., Edwards, H., Storkey, A., and Klimov, O. (2018).
\newblock Exploration by random network distillation.
\newblock {\em arXiv preprint arXiv:1810.12894}.

\bibitem[Carhart-Harris et~al., 2023]{carhart2023canalization}
Carhart-Harris, R., Chandaria, S., Erritzoe, D., Gazzaley, A., Girn, M., Kettner, H., Mediano, P., Nutt, D., Rosas, F., Roseman, L., et~al. (2023).
\newblock Canalization and plasticity in psychopathology.
\newblock {\em Neuropharmacology}, 226:109398.

\bibitem[Cobbe et~al., 2020]{cobbe2020leveraging}
Cobbe, K., Hesse, C., Hilton, J., and Schulman, J. (2020).
\newblock Leveraging procedural generation to benchmark reinforcement learning.
\newblock In {\em International conference on machine learning}, pages 2048--2056. PMLR.

\bibitem[Cobbe et~al., 2019]{cobbe2019quantifying}
Cobbe, K., Klimov, O., Hesse, C., Kim, T., and Schulman, J. (2019).
\newblock Quantifying generalization in reinforcement learning.
\newblock In {\em International Conference on Machine Learning}, pages 1282--1289. PMLR.

\bibitem[Dohare et~al., 2023a]{dohare2023loss}
Dohare, S., Hernandez-Garcia, J., Rahman, P., Sutton, R., and Mahmood, A.~R. (2023a).
\newblock Loss of plasticity in deep continual learning.

\bibitem[Dohare et~al., 2023b]{dohare2023maintaining}
Dohare, S., Hernandez-Garcia, J.~F., Rahman, P., Sutton, R.~S., and Mahmood, A.~R. (2023b).
\newblock Maintaining plasticity in deep continual learning.
\newblock {\em arXiv preprint arXiv:2306.13812}.

\bibitem[Dohare et~al., 2023c]{dohare2023overcoming}
Dohare, S., Lan, Q., and Mahmood, A.~R. (2023c).
\newblock Overcoming policy collapse in deep reinforcement learning.
\newblock In {\em Sixteenth European Workshop on Reinforcement Learning}.

\bibitem[D'Oro et~al., 2022]{d2022sample}
D'Oro, P., Schwarzer, M., Nikishin, E., Bacon, P.-L., Bellemare, M.~G., and Courville, A. (2022).
\newblock Sample-efficient reinforcement learning by breaking the replay ratio barrier.
\newblock In {\em Deep Reinforcement Learning Workshop NeurIPS 2022}.

\bibitem[Hensch, 2004]{hensch2004critical}
Hensch, T.~K. (2004).
\newblock Critical period regulation.
\newblock {\em Annu. Rev. Neurosci.}, 27:549--579.

\bibitem[Igl et~al., 2020]{igl2020transient}
Igl, M., Farquhar, G., Luketina, J., Boehmer, W., and Whiteson, S. (2020).
\newblock Transient non-stationarity and generalisation in deep reinforcement learning.
\newblock {\em arXiv preprint arXiv:2006.05826}.

\bibitem[Juliani et~al., 2022]{juliani2022neuro}
Juliani, A., Barnett, S., Davis, B., Sereno, M., and Momennejad, I. (2022).
\newblock Neuro-nav: a library for neurally-plausible reinforcement learning.
\newblock {\em arXiv preprint arXiv:2206.03312}.

\bibitem[Juliani et~al., 2024]{juliani2023deep}
Juliani, A., Safron, A., and Kanai, R. (2024).
\newblock Deep canals: A deep learning approach to refining the canalization theory of psychopathology.
\newblock {\em Neuroscience of Consciousness}.

\bibitem[Kumar et~al., 2023]{kumar2023maintaining}
Kumar, S., Marklund, H., and Van~Roy, B. (2023).
\newblock Maintaining plasticity via regenerative regularization.
\newblock {\em arXiv preprint arXiv:2308.11958}.

\bibitem[Lee et~al., 2024]{lee2024plastic}
Lee, H., Cho, H., Kim, H., Gwak, D., Kim, J., Choo, J., Yun, S.-Y., and Yun, C. (2024).
\newblock Plastic: Improving input and label plasticity for sample efficient reinforcement learning.
\newblock {\em Advances in Neural Information Processing Systems}, 36.

\bibitem[Lewandowski et~al., 2023]{lewandowski2023curvature}
Lewandowski, A., Tanaka, H., Schuurmans, D., and Machado, M.~C. (2023).
\newblock Curvature explains loss of plasticity.
\newblock {\em arXiv preprint arXiv:2312.00246}.

\bibitem[Lyle et~al., 2022]{lyle2022understanding}
Lyle, C., Rowland, M., and Dabney, W. (2022).
\newblock Understanding and preventing capacity loss in reinforcement learning.
\newblock {\em arXiv preprint arXiv:2204.09560}.

\bibitem[Lyle et~al., 2023]{lyle2023understanding}
Lyle, C., Zheng, Z., Nikishin, E., Pires, B.~A., Pascanu, R., and Dabney, W. (2023).
\newblock Understanding plasticity in neural networks.
\newblock {\em arXiv preprint arXiv:2303.01486}.

\bibitem[Nikishin et~al., 2023]{nikishin2023deep}
Nikishin, E., Oh, J., Ostrovski, G., Lyle, C., Pascanu, R., Dabney, W., and Barreto, A. (2023).
\newblock Deep reinforcement learning with plasticity injection.
\newblock {\em arXiv preprint arXiv:2305.15555}.

\bibitem[Nikishin et~al., 2022]{nikishin2022primacy}
Nikishin, E., Schwarzer, M., D’Oro, P., Bacon, P.-L., and Courville, A. (2022).
\newblock The primacy bias in deep reinforcement learning.
\newblock In {\em International conference on machine learning}, pages 16828--16847. PMLR.

\bibitem[Peled, 2005]{peled2005plasticity}
Peled, A. (2005).
\newblock Plasticity imbalance in mental disorders the neuroscience of psychiatry: implications for diagnosis and research.
\newblock {\em Medical hypotheses}, 65(5):947--952.

\bibitem[Salimans and Chen, 2018]{salimans2018learning}
Salimans, T. and Chen, R. (2018).
\newblock Learning montezuma's revenge from a single demonstration.
\newblock {\em arXiv preprint arXiv:1812.03381}.

\bibitem[Schaul et~al., 2019]{schaul2019ray}
Schaul, T., Borsa, D., Modayil, J., and Pascanu, R. (2019).
\newblock Ray interference: a source of plateaus in deep reinforcement learning.
\newblock {\em arXiv preprint arXiv:1904.11455}.

\bibitem[Schulman et~al., 2017]{schulman2017proximal}
Schulman, J., Wolski, F., Dhariwal, P., Radford, A., and Klimov, O. (2017).
\newblock Proximal policy optimization algorithms.
\newblock {\em arXiv preprint arXiv:1707.06347}.

\bibitem[Schwarzer et~al., 2023]{schwarzer2023bigger}
Schwarzer, M., Ceron, J. S.~O., Courville, A., Bellemare, M.~G., Agarwal, R., and Castro, P.~S. (2023).
\newblock Bigger, better, faster: Human-level atari with human-level efficiency.
\newblock In {\em International Conference on Machine Learning}, pages 30365--30380. PMLR.

\bibitem[Sokar et~al., 2023]{sokar2023dormant}
Sokar, G., Agarwal, R., Castro, P.~S., and Evci, U. (2023).
\newblock The dormant neuron phenomenon in deep reinforcement learning.
\newblock In {\em International Conference on Machine Learning}, pages 32145--32168. PMLR.

\bibitem[Wiesel and Hubel, 1963]{wiesel1963effects}
Wiesel, T.~N. and Hubel, D.~H. (1963).
\newblock Effects of visual deprivation on morphology and physiology of cells in the cat's lateral geniculate body.
\newblock {\em Journal of neurophysiology}, 26(6):978--993.

\bibitem[Xu et~al., 2023]{xu2023drm}
Xu, G., Zheng, R., Liang, Y., Wang, X., Yuan, Z., Ji, T., Luo, Y., Liu, X., Yuan, J., Hua, P., et~al. (2023).
\newblock Drm: Mastering visual reinforcement learning through dormant ratio minimization.
\newblock {\em arXiv preprint arXiv:2310.19668}.

\bibitem[Zaidi et~al., 2023]{zaidi2023does}
Zaidi, S., Berariu, T., Kim, H., Bornschein, J., Clopath, C., Teh, Y.~W., and Pascanu, R. (2023).
\newblock When does re-initialization work?
\newblock In {\em Proceedings on}, pages 12--26. PMLR.

\end{thebibliography}
\bibliographystyle{apalike}

\newpage
\clearpage
\appendix

\section{Implementation Details}\label{app:method_details}

Below are the specific implementation details for each intervention. Wherever possible we followed the methods provided by the papers which introduced each method. Relevant hyperparameters used are listed in Table \ref{tab:intervention_params}.

\textbf{Shrink+perturb.} When the intervention is applied all learnable parameters in the network are iterated through and scaled by $\alpha$. All parameters are then additively combined with newly sampled initialization parameters which are scaled by $\beta$. For a set of parameters $x$, this corresponds to $x_{new} = \alpha x_{current} + \beta x_{init}$, where $x_{init}$ is sampled from the parameter initialization distribution. For all experiments $\alpha = 1 - \beta$.

\textbf{Soft shrink+perturb.} The shrink+perturb procedure is applied after each step of gradient descent instead of only at specific intervals. Specifically, for a set of parameters $x$, this corresponds to $x_{new} = \alpha x_{current} + \beta x_{init}$, where $x_{init}$ is sampled from the parameter initialization distribution. For all experiments $\alpha = 1 - \beta$. 

\textbf{CReLU.} All ReLU activations in the network are replaced with CReLU. The equation for CReLU is: $y = concat(ReLU(x), ReLU(-x))$. We use the method where input parameter sizes are halved compared to the baseline so that output parameters remain consistent for each layer. This results in a network with half the number of parameters for the effected layers.

\textbf{Regenerative Regularization (L2).} An additional loss term is added to the PPO update in each step of gradient descent. This term corresponds to L2 norm of the difference between the current parameters of the network and the network parameters at initialization. This loss is scaled by $\alpha$. 

\textbf{Plasticity injection.} We follow the method described by \citeauthor{nikishin2023deep} for the initial intervention, only performing plasticity injection on the final layers of the network (i.e. value and policy heads). To ensure computational efficiency, during subsequent interventions we combine the previous parameter values into a single set of weights and biases (i.e. $w_{old} = w_{old} + w_{new a.} w_{new b}$). This ensures that at any given time the plasticity-injected layer is represented by $y = x_{old} + x_{new a.} sg(x_{new b})$.

\textbf{LayerNorm.} We apply the LayerNorm \cite{ba2016layer} function before the ReLU activation at each layer.

\textbf{Reset-final.} At each intervention point we replace the weights of the final layers (i.e. value and policy heads) of the network with freshly initialized values.

\textbf{L2 Norm.} We apply an additional loss term to the PPO update which corresponds to the L2 norm of the weights of the network. This loss is scaled by a parameter $\alpha$.

\textbf{ReDo.} At a fixed epoch interval the procedure from \citeauthor{sokar2023dormant} is performed on all layers of the neural network. A mini-batch of training data is used to calculate the dormancy rate for each unit, and units which are below $\tau$ are re-initialized.

\newpage

\section{Environment and Model Details}\label{app:env_details}

Additional hyperparameters used for Gridworld and CoinRun models are provided in Table \ref{tab:ppo_params}. All experiments involve running five repetitions with unique random seeds each. All experiments are conducted using either a single P100 or V100 GPU on a cloud machine. We used preliminary exploratory training runs to determine the necessary number of epochs to ensure convergence within the initial single round of training. In these longer training runs we did not find evidence of policy collapse \cite{dohare2023overcoming}, and thus believe that performance degradation in subsequent rounds can be attributed to plasticity loss. Code which can be used to reproduce our results is available at \href{https://github.com/awjuliani/deep-rl-plasticity}{https://github.com/awjuliani/deep-rl-plasticity}.

\textbf{Gridworld.} The gridworld environment consisted of an 11x11x4 one-hot encoded observation space. Each observation was flattened into a single vector and provided as input to the model. In the permute environment variation, observations were split into 1x1x4 patches and randomized according to a fixed map within each round. The PPO model consisted of an MLP with two hidden layers, ReLU activations, and a ``dual-head'' architecture, as described in \cite{schulman2017proximal}. The action space consisted of four possible actions, corresponding to movement in the cardinal directions. All episodes lasted 100 time-steps. Each round of training consists of 20,000 iterations of inference and model updates. Given ten rounds, the total number of iterations is 200,000.

\textbf{CoinRun.} The CoinRun environment consisted of a 64x64x3 observation space. This was provided to the PPO model as a tensor, and processed by a convolutional encoder with four layers and ReLU activations. Each convolutional layer utilized a kernel size of four and a stride of two. In the permute environment variation, images were broken into 8x8x3 patches and the position of patches was randomized according to a fixed map within each round. A ``dual-head'' architecture was also utilized. The action space consisted of the standard nine possible actions provided by the ProcGen environment~\cite{cobbe2020leveraging}. Episode length varied based on agent behavior. Each round of training consists of 5,000 iterations of inference and model updates. Given ten rounds, the total number of iterations is 50,000.

\textbf{Montezuma's Revenge.} Like in CoinRun, Montezuma's revenge states are processed into observations of size 64x64x3 and passed through a convolutional encoder. We inherit hyperparameters from~\citet{burda2018exploration}, which were previously optimized for final performance. These include 4 randomly masked update epochs per observation, an entropy coefficient of 0.001, $\gamma_E$ = 0.999 and $\gamma_I$ =0.99. We use a learning rate of 0.0001 and 128 simultaneous agents simultaneously. All model updates are performed via the Adam optimizer. We use a shrinkage coefficient $1-10^{-8}$ and a perturbation coefficient of $10^{-8}$.

\newpage

\section{Hyperparameters}

\begin{table}[!th]
    \centering
    \begin{tabular}{l|ll}
         \textbf{Method}&  \textbf{Gridworld}& \textbf{CoinRun}\\
         \hline
         L2-Norm ($\alpha$)&  1e-3& 1e-3\\
         Regen-Reg ($\alpha$)& 1e-4& 1e-2\\
         Soft-S+P ($\beta$)&  1e-6& 1e-6\\
         S+P ($\beta$)& 0.5&0.5\\
         ReDo period & 10 epochs & 10 epochs\\
         ReDo ($\tau$) & 0.025 & 0.025\\
    \end{tabular}
    \caption{Optimal hyperparameter values used for different environments and interventions. Chosen values are the result of performing a sweep over possible values using \textit{permute} environment shift condition.}
    \label{tab:intervention_params}
\end{table}

\begin{table}[!th]
    \centering
    \begin{tabular}{l|l}
         \textbf{Parameter}& \textbf{Value}\\
         \hline
         Number of hidden units& 256\\
         Learning rate& 5e-4\\
         Default activation& ReLU\\
         Minibatch size & 64\\
         Buffer size& 1024\\
         Discount ($\gamma$)& 0.99\\
         GAE parameter ($\lambda$)&0.95\\
         Clip parameter ($\epsilon$)&0.2\\
         Entropy parameter ($\beta$)&0.02\\
         Number of update epochs&3\\
    \end{tabular}
    \caption{Default values used in PPO algorithm for all gridworld and CoinRun experiments.}
    \label{tab:ppo_params}
\end{table}

\newpage

\section{Extended Results}\label{app:results}

\begin{figure*}[h]
    \centering
    \includegraphics[width=1.0\linewidth]{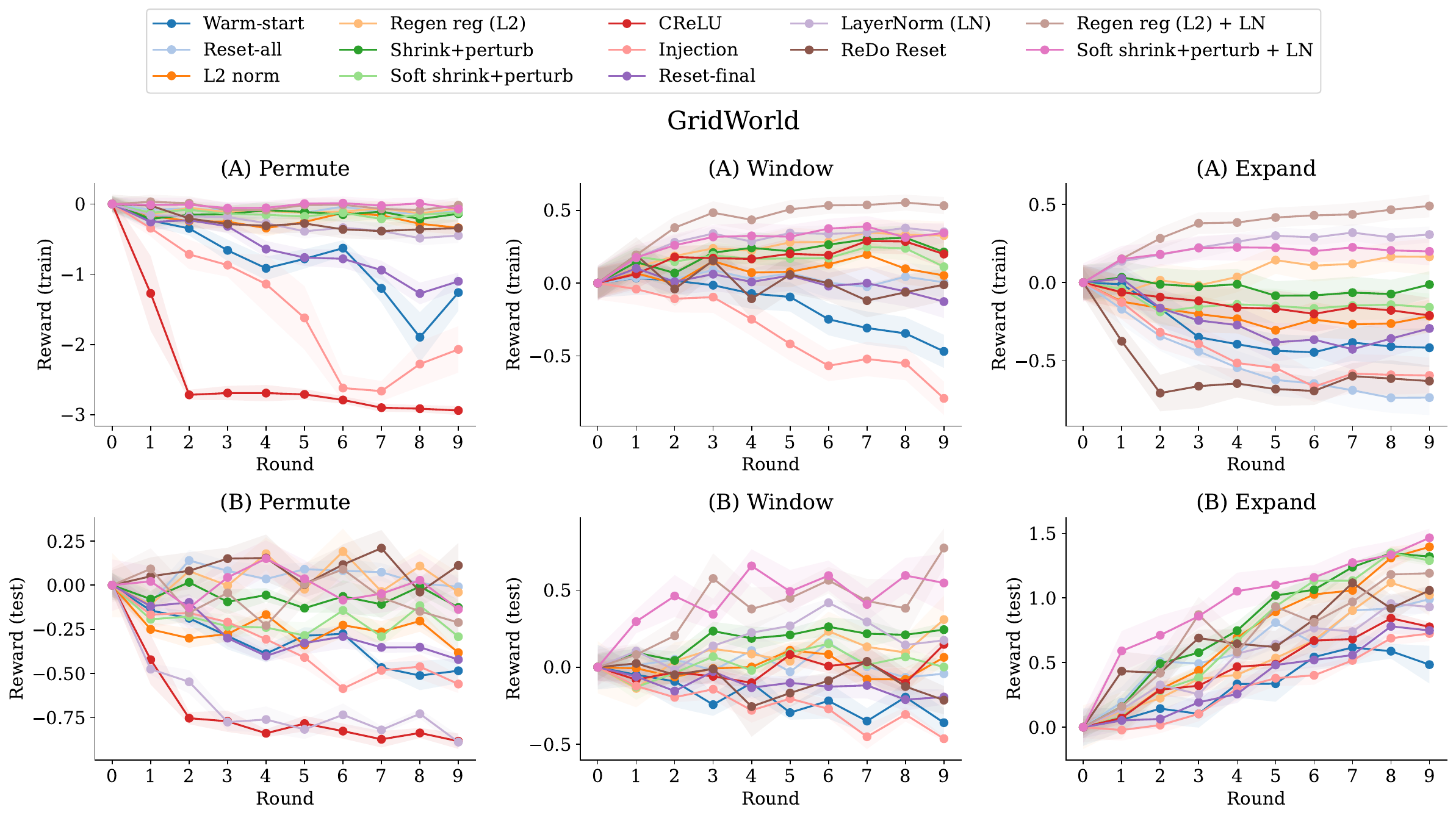}
    \caption{Performance of intervention methods compared to baseline (warm-start) on Gridworld task. Shaded region corresponds to normalized mean episodic reward. \textbf{Tow row}: Training distribution performance. \textbf{Bottom row}: Test distribution performance.}
    \label{fig:gridworld_full}
\end{figure*}

\begin{figure*}[h]
    \centering
    \includegraphics[width=1.0\linewidth]{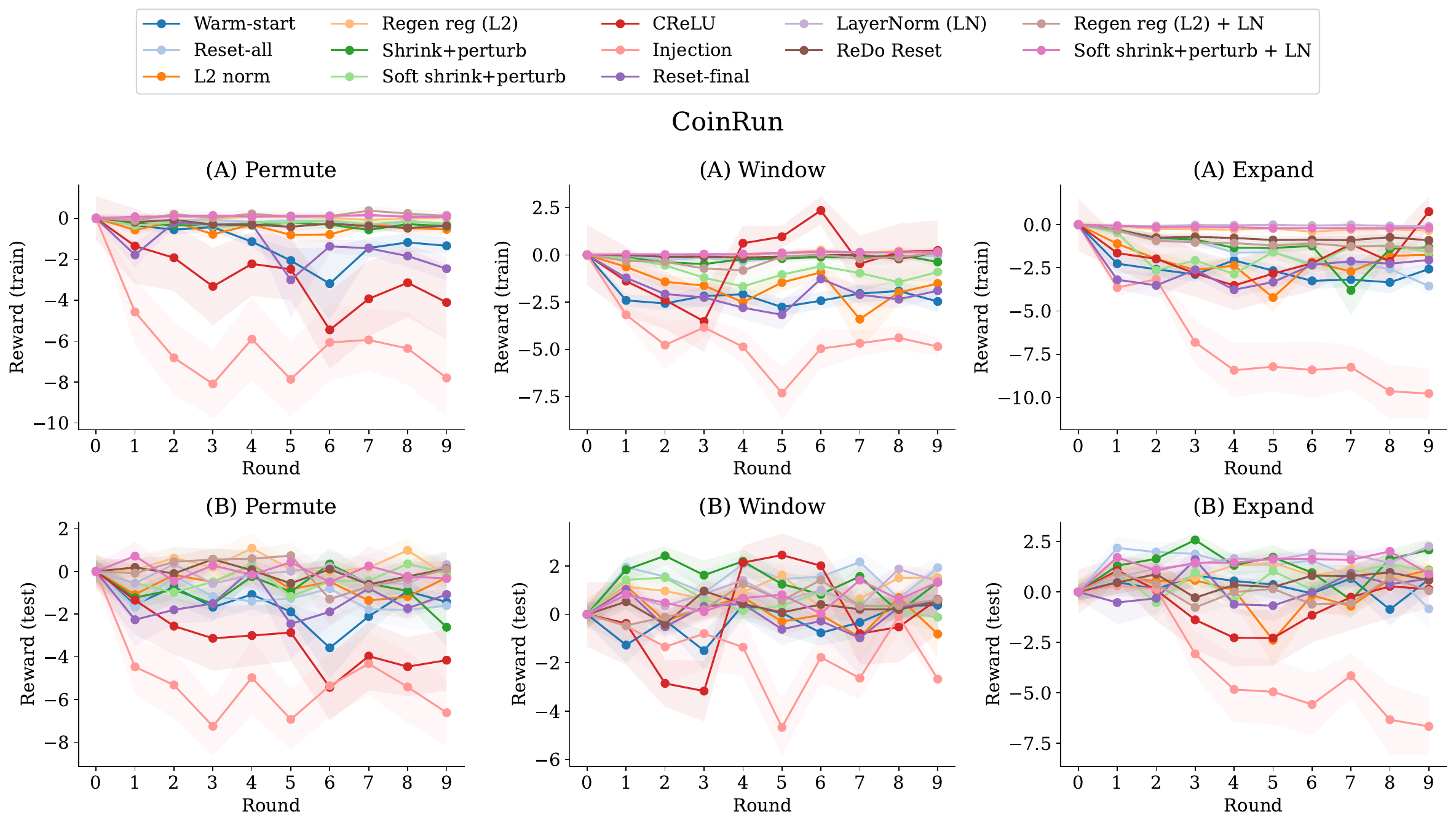}
    \caption{Performance of intervention methods compared to baseline (warm-start) on CoinRun task. Shaded region corresponds to normalized mean episodic reward. \textbf{Top row}: Training distribution performance. \textbf{Bottom row}: Test distribution performance.}
    \label{fig:coinrun_full}
\end{figure*}

\begin{figure*}[!h]
    \centering
    \includegraphics[width=1.0\linewidth]{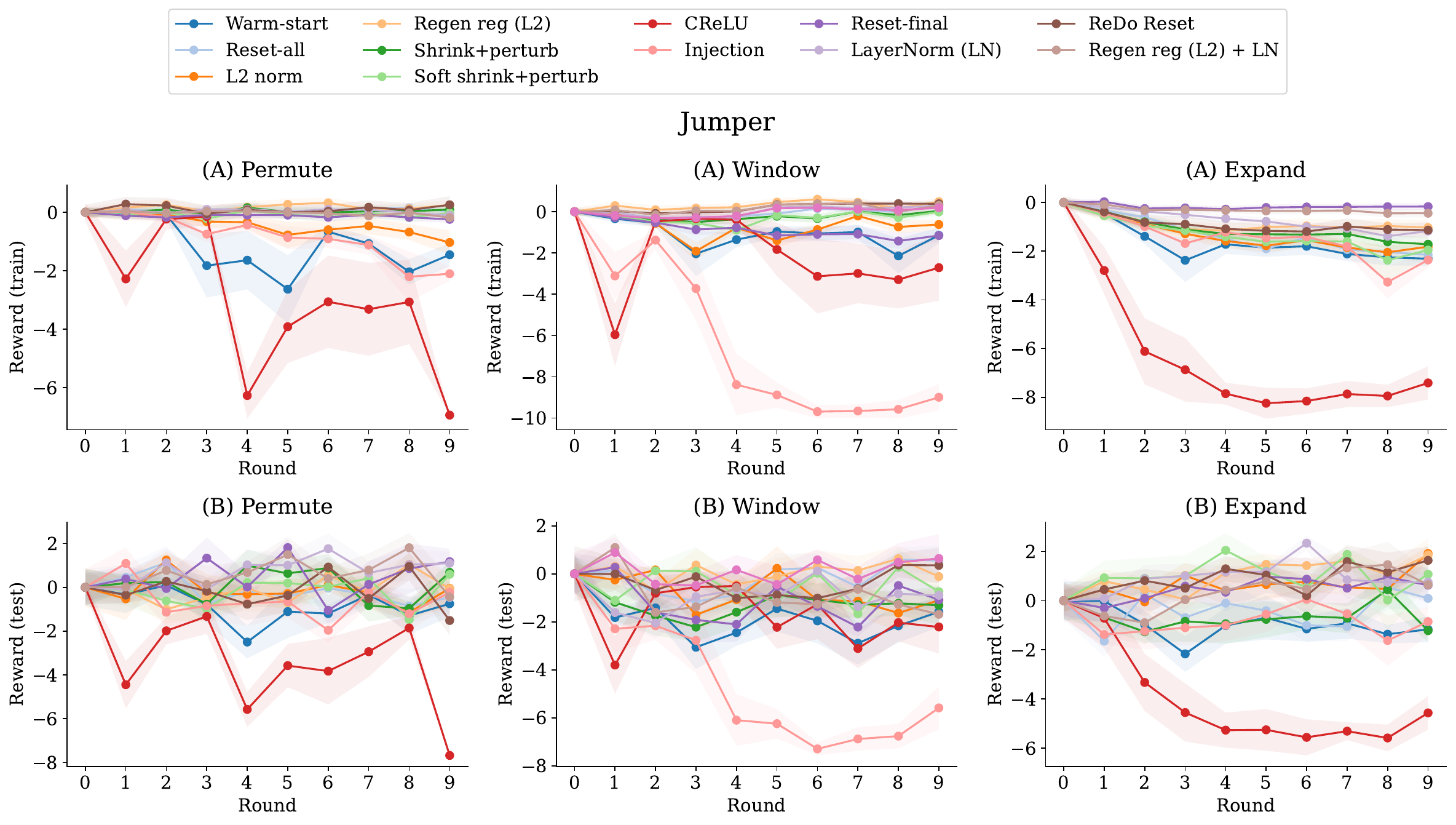}
    \caption{Performance of intervention methods compared to baseline (warm-start) on Jumper task. \textbf{Top row}: Training distribution performance. \textbf{Bottom row}: Test distribution performance.}
    \label{fig:jumper_full}
\end{figure*}

\begin{figure*}[!h]
    \centering
    \includegraphics[width=1.0\linewidth]{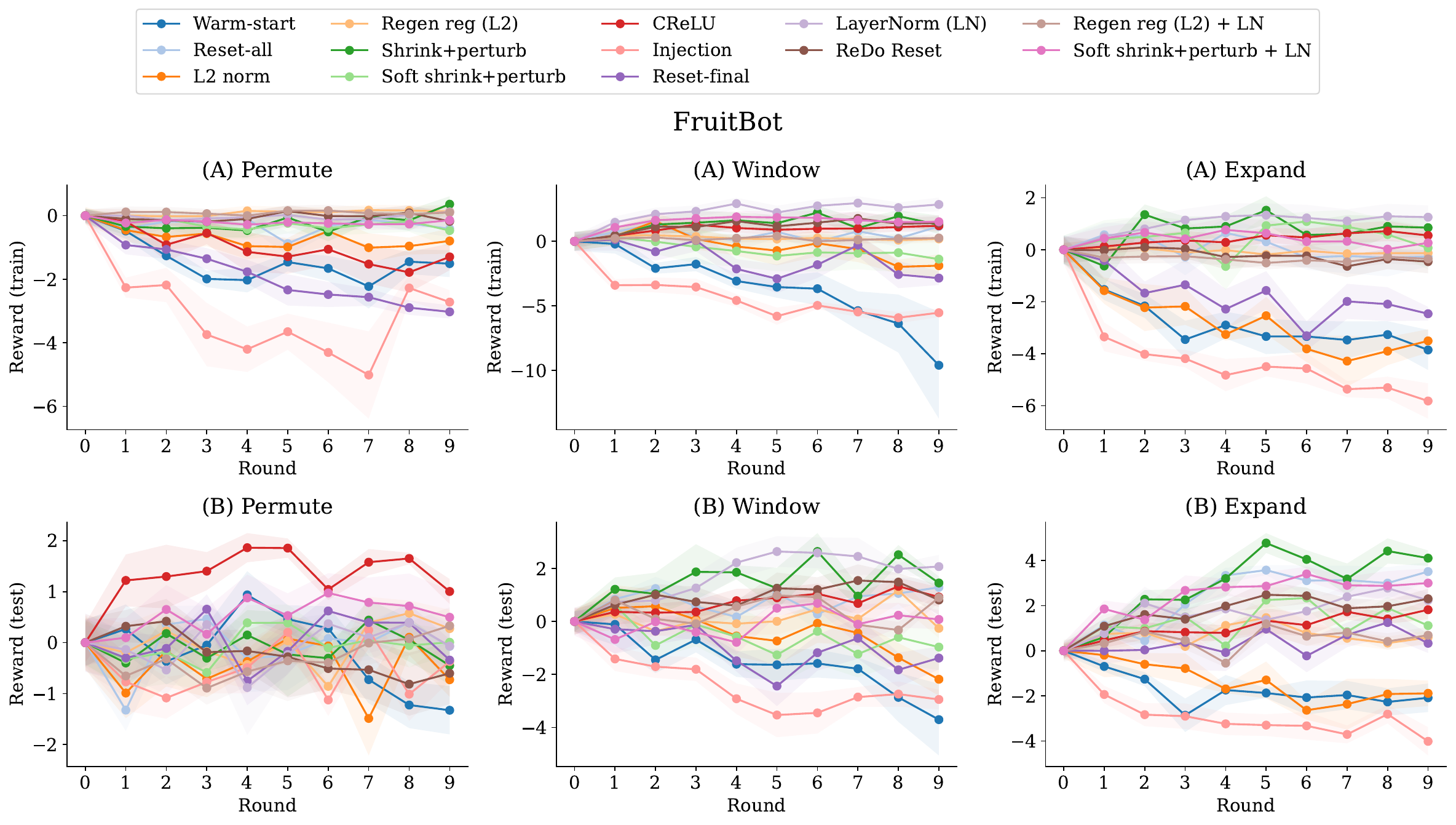}
    \caption{Performance of intervention methods compared to baseline (warm-start) on Fruitbot task. \textbf{Top row}: Training distribution performance. \textbf{Bottom row}: Test distribution performance.}
    \label{fig:fruitbot_full}
\end{figure*}

\newpage

\begin{figure*}[!h]
    \centering
    \includegraphics[width=1.0\linewidth]{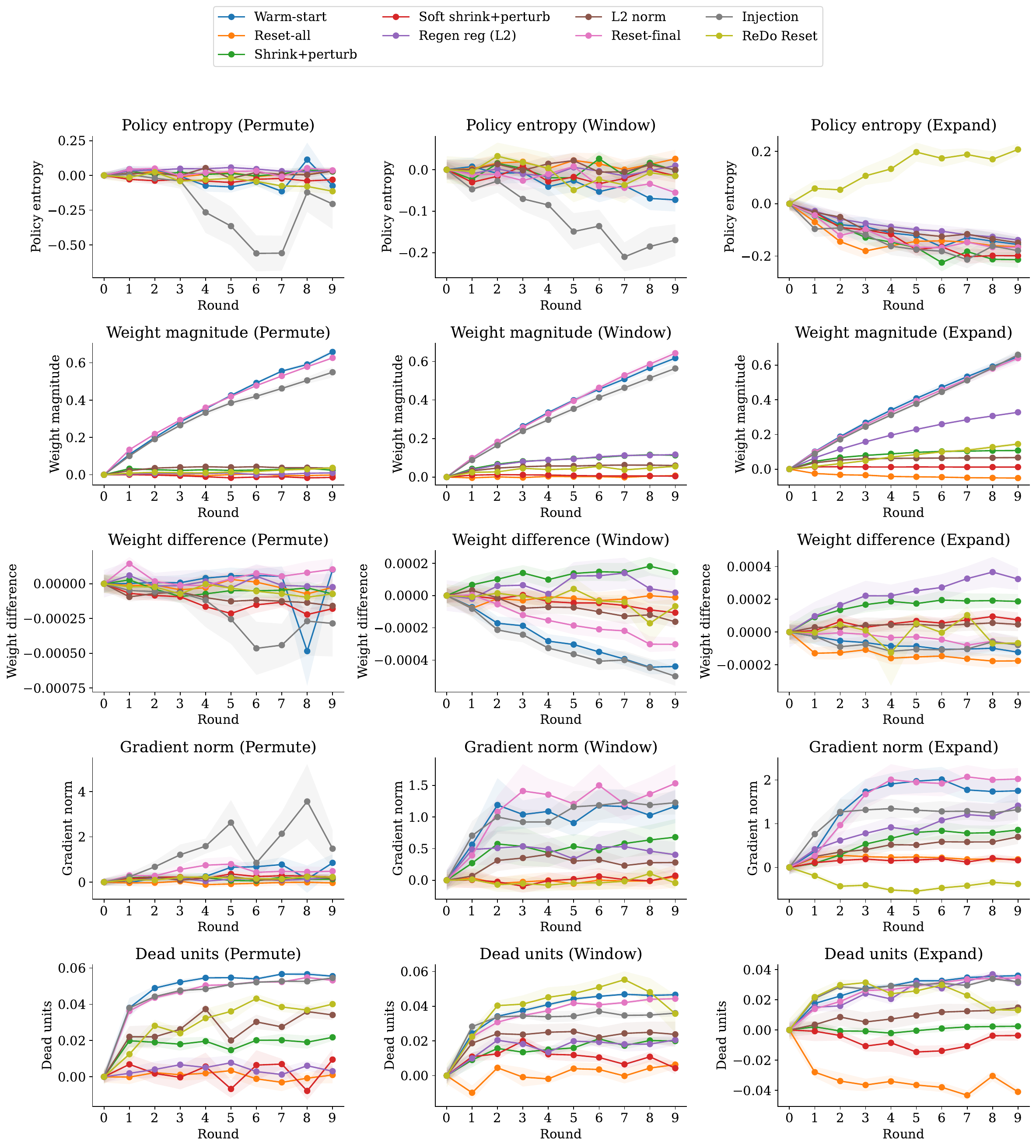}
    \caption{Effect of learning under various conditions on five diagnostic metrics of interest.}
    \label{fig:extra_metrics_full}
\end{figure*}

\newpage

\begin{table}[h]
\begin{center}
\begin{tabular}{lclc}
\toprule
\textbf{Dep. Variable:}   &        train\_reward         & \textbf{  No. Observations:  } &       27    \\
\textbf{Model:}           &       GLM        & \textbf{  Df Residuals:      } &       21    \\
\textbf{Model Family:}    &     Gaussian     & \textbf{  Df Model:          } &        5    \\
\textbf{Link Function:}   &     Identity     & \textbf{  Scale:             } &   0.13264   \\
\textbf{Method:}          &       IRLS       & \textbf{  Log-Likelihood:    } &   -7.6473   \\
            &  & \textbf{  Deviance:          } &    2.7855   \\
            &          & \textbf{  Pearson chi2:      } &     2.79    \\
\textbf{No. Iterations:}  &        3         & \textbf{  Pseudo R-squ. (CS):} &   0.5624    \\
\bottomrule
\end{tabular}
\begin{tabular}{lcccccc}
               & \textbf{coef} & \textbf{std err} & \textbf{z} & \textbf{P$> |$z$|$} & \textbf{[0.025} & \textbf{0.975]}  \\
\midrule
\textbf{const} &      -0.0920  &        0.118     &    -0.783  &         0.434        &       -0.322    &        0.138     \\
\textbf{entropy}    &       1.6302  &        0.940     &     1.734  &         0.083        &       -0.213    &        3.473     \\
\textbf{weight\_mag}    &      -1.8262  &        0.637     &    -2.868  &         0.004        &       -3.074    &       -0.578     \\
\textbf{weight\_diff}    &     -35.4734  &      133.365     &    -0.266  &         0.790        &     -296.865    &      225.918     \\
\textbf{grad\_norm}    &       0.5742  &        0.184     &     3.116  &         0.002        &        0.213    &        0.935     \\
\textbf{dead\_units}    &       0.5074  &        5.905     &     0.086  &         0.932        &      -11.066    &       12.081     \\
\bottomrule
\end{tabular}
\end{center}
\caption{Results of GLM using various diagnostic metrics to predict reward during training.}
\end{table}

\begin{table}[h]
\begin{center}
\begin{tabular}{lclc}
\toprule
\textbf{Dep. Variable:}   &        test\_reward         & \textbf{  No. Observations:  } &       27    \\
\textbf{Model:}           &       GLM        & \textbf{  Df Residuals:      } &       21    \\
\textbf{Model Family:}    &     Gaussian     & \textbf{  Df Model:          } &        5    \\
\textbf{Link Function:}   &     Identity     & \textbf{  Scale:             } &   0.19798   \\
\textbf{Method:}          &       IRLS       & \textbf{  Log-Likelihood:    } &   -13.054   \\
            &  & \textbf{  Deviance:          } &    4.1576   \\
            &          & \textbf{  Pearson chi2:      } &     4.16    \\
\textbf{No. Iterations:}  &        3         & \textbf{  Pseudo R-squ. (CS):} &   0.5215    \\
\bottomrule
\end{tabular}
\begin{tabular}{lcccccc}
               & \textbf{coef} & \textbf{std err} & \textbf{z} & \textbf{P$> |$z$|$} & \textbf{[0.025} & \textbf{0.975]}  \\
\midrule
\textbf{const} &       0.3249  &        0.144     &     2.263  &         0.024        &        0.043    &        0.606     \\
\textbf{entropy}    &      -1.4060  &        1.149     &    -1.224  &         0.221        &       -3.657    &        0.845     \\
\textbf{weight\_mag}    &       1.1259  &        0.778     &     1.447  &         0.148        &       -0.399    &        2.651     \\
\textbf{weight\_diff}    &     327.8923  &      162.934     &     2.012  &         0.044        &        8.547    &      647.238     \\
\textbf{grad\_norm}    &      -0.0568  &        0.225     &    -0.252  &         0.801        &       -0.498    &        0.384     \\
\textbf{dead\_units}    &     -21.6282  &        7.214     &    -2.998  &         0.003        &      -35.768    &       -7.488     \\
\bottomrule
\end{tabular}
\end{center}
\caption{Results of GLM using various diagnostic metrics to predict reward on the test distribution.}
\end{table}

\begin{table}[!ht]
    \centering
    \begin{tabular}{|l|l|l|}%
    \hline
    \multicolumn{3}{|c|}{\textbf{Gridworld (Permute - Train)}}\\
    \hline%
    \textbf{Method}&\textbf{t{-}value}&\textbf{p{-}value}\\%
    \hline%
    Warm{-}start&t(7) = 2.524&0.04\\%
    \hline%
    CReLU&t(7) = 10.477&0.0\\%
    \hline%
    Injection&t(8) = 6.21&0.0\\%
    \hline%
    Reset{-}final&t(7) = 2.444&0.044\\%
    \hline%
    L2 norm&t(7) = {-}0.08&0.939\\%
    \hline%
    Shrink+perturb&t(8) = {-}0.087&0.933\\%
    \hline%
    Soft shrink+perturb&t(8) = 0.472&0.65\\%
    \hline%
    Regen reg (L2)&t(8) = 0.134&0.897\\%
    \hline%
    LayerNorm (LN)&t(8) = 0.787&0.454\\%
    \hline%
    Soft shrink+perturb + LN&t(8) = {-}0.212&0.837\\%
    \hline%
    Regen reg (L2) + LN&t(8) = {-}1.368&0.208\\%
    \hline%
    \end{tabular}%
    \caption{Results of statistical tests comparing different methods to baseline of \textit{reset-all} in gridworld (permute) condition. T-values below zero correspond to train distribution performance better than baseline. T-values above zero correspond to performance worse than baseline. P-values above 0.05 correspond to methods which produce mean normalized rewards not significantly different from the baseline.}
    \label{tab:grid_permute_stats}
\end{table}

\begin{table}[!ht]
    \centering
    \begin{tabular}{|l|l|l|}%
    \hline
    \multicolumn{3}{|c|}{\textbf{Gridworld (Window - Train)}}\\
    \hline%
    \textbf{Method}&\textbf{t{-}value}&\textbf{p{-}value}\\%
    \hline%
    Warm{-}start&t(8) = 0.759&0.469\\%
    \hline%
    CReLU&t(7) = {-}1.563&0.162\\%
    \hline%
    Injection&t(8) = 1.741&0.12\\%
    \hline%
    Reset{-}final&t(8) = {-}0.017&0.987\\%
    \hline%
    L2 norm&t(8) = {-}1.291&0.233\\%
    \hline%
    Shrink+perturb&t(8) = {-}1.361&0.211\\%
    \hline%
    Soft shrink+perturb&t(7) = {-}0.222&0.831\\%
    \hline%
    Regen reg (L2)&t(8) = {-}1.068&0.317\\%
    \hline%
    LayerNorm (LN)&t(7) = {-}2.107&0.073\\%
    \hline%
    Soft shrink+perturb + LN&t(8) = {-}1.549&0.16\\%
    \hline%
    Regen reg (L2) + LN&t(8) = {-}3.106&0.015\\%
    \hline%
    \end{tabular}%
    \caption{Results of statistical tests comparing different methods to baseline of \textit{reset-all} in gridworld (window) condition. T-values below zero correspond to train distribution performance better than baseline. T-values above zero correspond to performance worse than baseline. P-values above 0.05 correspond to methods which produce mean normalized rewards not significantly different from the baseline.}
    \label{tab:grid_window_stats}
\end{table}

\begin{table}[!ht]
    \centering
    \begin{tabular}{|l|l|l|}%
    \hline
    \multicolumn{3}{|c|}{\textbf{Gridworld (Expand - Train)}}\\
    \hline%
    \textbf{Method}&\textbf{t{-}value}&\textbf{p{-}value}\\%
    \hline%
    Warm{-}start&t(8) = {-}0.926&0.382\\%
    \hline%
    CReLU&t(8) = {-}2.247&0.055\\%
    \hline%
    Injection&t(8) = {-}0.715&0.495\\%
    \hline%
    Reset{-}final&t(8) = {-}1.301&0.229\\%
    \hline%
    L2 norm&t(8) = {-}2.252&0.054\\%
    \hline%
    Shrink+perturb&t(8) = {-}2.476&0.038\\%
    \hline%
    Soft shrink+perturb&t(7) = {-}1.788&0.117\\%
    \hline%
    Regen reg (L2)&t(8) = {-}2.879&0.021\\%
    \hline%
    LayerNorm (LN)&t(8) = {-}4.098&0.003\\%
    \hline%
    Soft shrink+perturb + LN&t(8) = {-}3.683&0.006\\%
    \hline%
    Regen reg (L2) + LN&t(8) = {-}5.44&0.001\\%
    \hline%
    \end{tabular}%
    \caption{Results of statistical tests comparing different methods to baseline of \textit{reset-all} in gridworld (expand) condition. T-values below zero correspond to train distribution performance better than baseline. T-values above zero correspond to performance worse than baseline. P-values above 0.05 correspond to methods which produce mean normalized rewards not significantly different from the baseline.}
    \label{tab:grid_expand_stats}
\end{table}

\begin{table}[!ht]
    \centering
    \begin{tabular}{|l|l|l|}%
    \hline
    \multicolumn{3}{|c|}{\textbf{Gridworld (Permute - Test)}}\\
    \hline%
    \textbf{Method}&\textbf{t{-}value}&\textbf{p{-}value}\\%
    \hline%
    Warm{-}start&t(7) = 3.0&0.02\\%
    \hline%
    CReLU&t(7) = 6.975&0.0\\%
    \hline%
    Injection&t(8) = 3.595&0.007\\%
    \hline%
    Reset{-}final&t(7) = 2.445&0.044\\%
    \hline%
    L2 norm&t(7) = 3.009&0.02\\%
    \hline%
    Shrink+perturb&t(8) = 0.907&0.391\\%
    \hline%
    Soft shrink+perturb&t(8) = 2.177&0.061\\%
    \hline%
    Regen reg (L2)&t(8) = 0.007&0.995\\%
    \hline%
    LayerNorm (LN)&t(8) = 6.959&0.0\\%
    \hline%
    Soft shrink+perturb + LN&t(8) = 0.235&0.82\\%
    \hline%
    Regen reg (L2) + LN&t(8) = 0.804&0.445\\%
    \hline%
    \end{tabular}%
    \caption{Results of statistical tests comparing different methods to baseline of \textit{reset-all} in gridworld (permute) condition. T-values below zero correspond to test distribution performance better than baseline. T-values above zero correspond to performance worse than baseline. P-values above 0.05 correspond to methods which produce mean normalized rewards not significantly different from the baseline.}
    \label{tab:grid_permute_test_stats}
\end{table}

\begin{table}[!ht]
    \centering
    \begin{tabular}{|l|l|l|}%
    \hline
    \multicolumn{3}{|c|}{\textbf{Gridworld (Window - Test)}}\\
    \hline%
    \textbf{Method}&\textbf{t{-}value}&\textbf{p{-}value}\\%
    \hline%
    Warm{-}start&t(8) = 0.882&0.404\\%
    \hline%
    CReLU&t(7) = 0.239&0.818\\%
    \hline%
    Injection&t(8) = 1.444&0.187\\%
    \hline%
    Reset{-}final&t(8) = 0.804&0.445\\%
    \hline%
    L2 norm&t(8) = 0.216&0.834\\%
    \hline%
    Shrink+perturb&t(8) = {-}0.781&0.458\\%
    \hline%
    Soft shrink+perturb&t(7) = {-}0.046&0.965\\%
    \hline%
    Regen reg (L2)&t(8) = {-}0.187&0.856\\%
    \hline%
    LayerNorm (LN)&t(7) = {-}0.813&0.443\\%
    \hline%
    Soft shrink+perturb + LN&t(8) = {-}2.349&0.047\\%
    \hline%
    Regen reg (L2) + LN&t(8) = {-}1.714&0.125\\%
    \hline%
    \end{tabular}%
    \caption{Results of statistical tests comparing different methods to baseline of \textit{reset-all} in gridworld (window) condition. T-values below zero correspond to test distribution performance better than baseline. T-values above zero correspond to performance worse than baseline. P-values above 0.05 correspond to methods which produce mean normalized rewards not significantly different from the baseline.}
    \label{tab:grid_window_test_stats}
\end{table}

\begin{table}[!ht]
    \centering
    \begin{tabular}{|l|l|l|}%
    \hline
    \multicolumn{3}{|c|}{\textbf{Gridworld (Expand - Test)}}\\
    \hline%
    \textbf{Method}&\textbf{t{-}value}&\textbf{p{-}value}\\%
    \hline%
    Warm{-}start&t(8) = 1.227&0.255\\%
    \hline%
    CReLU&t(8) = 0.891&0.399\\%
    \hline%
    Injection&t(8) = 1.815&0.107\\%
    \hline%
    Reset{-}final&t(8) = 1.457&0.183\\%
    \hline%
    L2 norm&t(8) = {-}0.512&0.623\\%
    \hline%
    Shrink+perturb&t(8) = {-}1.021&0.337\\%
    \hline%
    Soft shrink+perturb&t(7) = {-}0.527&0.614\\%
    \hline%
    Regen reg (L2)&t(8) = 0.333&0.748\\%
    \hline%
    LayerNorm (LN)&t(8) = 0.438&0.673\\%
    \hline%
    Soft shrink+perturb + LN&t(8) = {-}1.793&0.111\\%
    \hline%
    Regen reg (L2) + LN&t(8) = {-}0.579&0.579\\%
    \hline%
        \end{tabular}%
    \caption{Results of statistical tests comparing different methods to baseline of \textit{reset-all} in gridworld (expand) condition. T-values below zero correspond to test distribution performance better than baseline. T-values above zero correspond to performance worse than baseline. P-values above 0.05 correspond to methods which produce mean normalized rewards not significantly different from the baseline.}
    \label{tab:grid_expand_test_stats}
\end{table}

\end{document}